\documentclass[10pt,twocolumn,letterpaper]{article}

\usepackage{cvpr}              

\usepackage{graphicx}
\usepackage{amsmath}
\usepackage{amssymb}
\usepackage{booktabs}
\usepackage{multicol}
\usepackage{multirow}
\usepackage{soul}
\usepackage{threeparttable} 
\usepackage[pagebackref=true,breaklinks=true,colorlinks,bookmarks=false]{hyperref}

\usepackage[capitalize]{cleveref}
\crefname{section}{Sec.}{Secs.}
\Crefname{section}{Section}{Sections}
\Crefname{table}{Table}{Tables}
\crefname{table}{Tab.}{Tabs.}

\def\cvprPaperID{7274} 
\def\confName{CVPR}
\def\confYear{2023}

\usepackage{cases}
\usepackage{bm}
\usepackage{cases}
\usepackage{multirow}
\usepackage{color} 
\usepackage{enumerate}
\usepackage{etoolbox}
\usepackage{mathrsfs}
\usepackage{subcaption}

\newcommand{\MyMapTemplatePrefix}[4]{\expandafter#1\csname#3#4\endcsname{#2{#4}}}
\newcommand{\MyMapTemplatePrefixNew}[5]{\expandafter#1\csname#4#5\endcsname{#2{#3{#5}}}}
\forcsvlist{\MyMapTemplatePrefix {\def} {\mathbf} {}} {A,B,C,D,E,F,G,H,I,J,K,L,M,N,O,P,Q,R,S,T,U,V,W,X,Y,Z}
\forcsvlist{\MyMapTemplatePrefix {\def} {\mathbf} {}} {a,b,c,d,e,f,g,h,i,j,k,l,m,n,o,p,q,r,s,t,u,v,w,x,y,z,1,0}
\forcsvlist{\MyMapTemplatePrefix {\def} {\widetilde} {wt}} {A,B,C,D,E,F,G,H,I,J,K,L,M,N,O,P,Q,R,S,T,U,V,W,X,Y,Z}
\forcsvlist{\MyMapTemplatePrefix {\def} {\widetilde} {wt}} {a,b,c,d,e,f,g,h,i,j,k,l,m,n,o,p,q,r,s,t,u,v,w,x,y,z} 
\forcsvlist{\MyMapTemplatePrefixNew {\def} {\widetilde}{\mathbf} {tb}} {A,B,C,D,E,F,G,H,I,J,K,L,M,N,O,P,Q,R,S,T,U,V,W,X,Y,Z}
\forcsvlist{\MyMapTemplatePrefixNew {\def} {\widetilde}{\mathbf} {tb}} {a,b,c,d,e,f,g,h,i,j,k,l,m,n,o,p,q,r,s,t,u,v,w,x,y,z}
\forcsvlist{\MyMapTemplatePrefix {\def} {\widehat} {wh}} {A,B,C,D,E,F,G,H,I,J,K,L,M,N,O,P,Q,R,S,T,U,V,W,X,Y,Z}
\forcsvlist{\MyMapTemplatePrefix {\def} {\widehat} {wh}} {a,b,c,d,e,f,g,h,i,j,k,l,m,n,o,p,q,r,s,t,u,v,w,x,y,z}
\forcsvlist{\MyMapTemplatePrefixNew {\def} {\widehat}{\mathbf} {hb}} {A,B,C,D,E,F,G,H,I,J,K,L,M,N,O,P,Q,R,S,T,U,V,W,X,Y,Z}
\forcsvlist{\MyMapTemplatePrefixNew {\def} {\widehat}{\mathbf} {hb}} {a,b,c,d,e,f,g,h,i,j,k,l,m,n,o,p,q,r,s,t,u,v,w,x,y,z}
\forcsvlist{\MyMapTemplatePrefixNew {\def} {\overline}{\mathbf} {lb}} {A,B,C,D,E,F,G,H,I,J,K,L,M,N,O,P,Q,R,S,T,U,V,W,X,Y,Z}
\forcsvlist{\MyMapTemplatePrefixNew {\def} {\overline}{\mathbf} {lb}} {a,b,c,d,e,f,g,h,i,j,k,l,m,n,o,p,q,r,s,t,u,v,w,x,y,z}
\forcsvlist{\MyMapTemplatePrefix {\def} {\mathcal}{mc}} {A,B,C,D,E,F,G,H,I,J,K,L,M,N,O,P,Q,R,S,T,U,V,W,X,Y,Z}
\forcsvlist{\MyMapTemplatePrefix {\def} {\mathbb} {mb}} {A,B,C,D,E,F,G,H,I,J,K,L,M,N,O,P,Q,R,S,T,U,V,W,X,Y,Z}
\forcsvlist{\MyMapTemplatePrefix {\DeclareMathOperator} {} {} } {tr,diag,sgn}

\def\sha{\text{com}}
\def\prt{\text{prt}}

\allowdisplaybreaks[4]

\begin{document}

\title{Decoupled Multimodal Distilling for Emotion Recognition}

\author{Yong Li, Yuanzhi Wang, Zhen Cui$^{*}$\\
	PCA Lab, Key Lab of Intelligent Perception and Systems for High-Dimensional\\
	Information of Ministry of Education, School of Computer Science and Engineering,\\
    Nanjing University of Science and Technology, Nanjing, China.\\
{\tt\small \{yong.li, yuanzhiwang, zhen.cui\}@njust.edu.cn}
}
\maketitle

\begin{abstract}
Human multimodal emotion recognition (MER) aims to perceive human emotions via language, visual and acoustic modalities. Despite the impressive performance of previous MER approaches, the inherent multimodal heterogeneities still haunt and the contribution of different modalities varies significantly. In this work, we mitigate this issue by proposing a decoupled multimodal distillation (DMD) approach that facilitates flexible and adaptive crossmodal knowledge distillation, aiming to enhance the discriminative features of each modality. Specially, the representation of each modality is decoupled into two parts, i.e., modality-irrelevant/-exclusive spaces, in a self-regression manner. DMD utilizes a graph distillation unit (GD-Unit) for each decoupled part so that each GD can be performed in a more specialized and effective manner. A GD-Unit consists of a dynamic graph where each vertice represents a modality and each edge indicates a dynamic knowledge distillation. Such GD paradigm provides a flexible knowledge transfer manner where the distillation weights can be automatically learned, thus enabling diverse crossmodal knowledge transfer patterns. Experimental results show DMD consistently obtains superior performance than state-of-the-art MER methods. Visualization results show the graph edges in DMD exhibit meaningful distributional patterns w.r.t. the modality-irrelevant/-exclusive feature spaces.
Codes are released at \url{https://github.com/mdswyz/DMD}.
\end{abstract}
\vspace{-0.6cm}


\begin{figure}[htb]
	\includegraphics[width=1.0\linewidth]{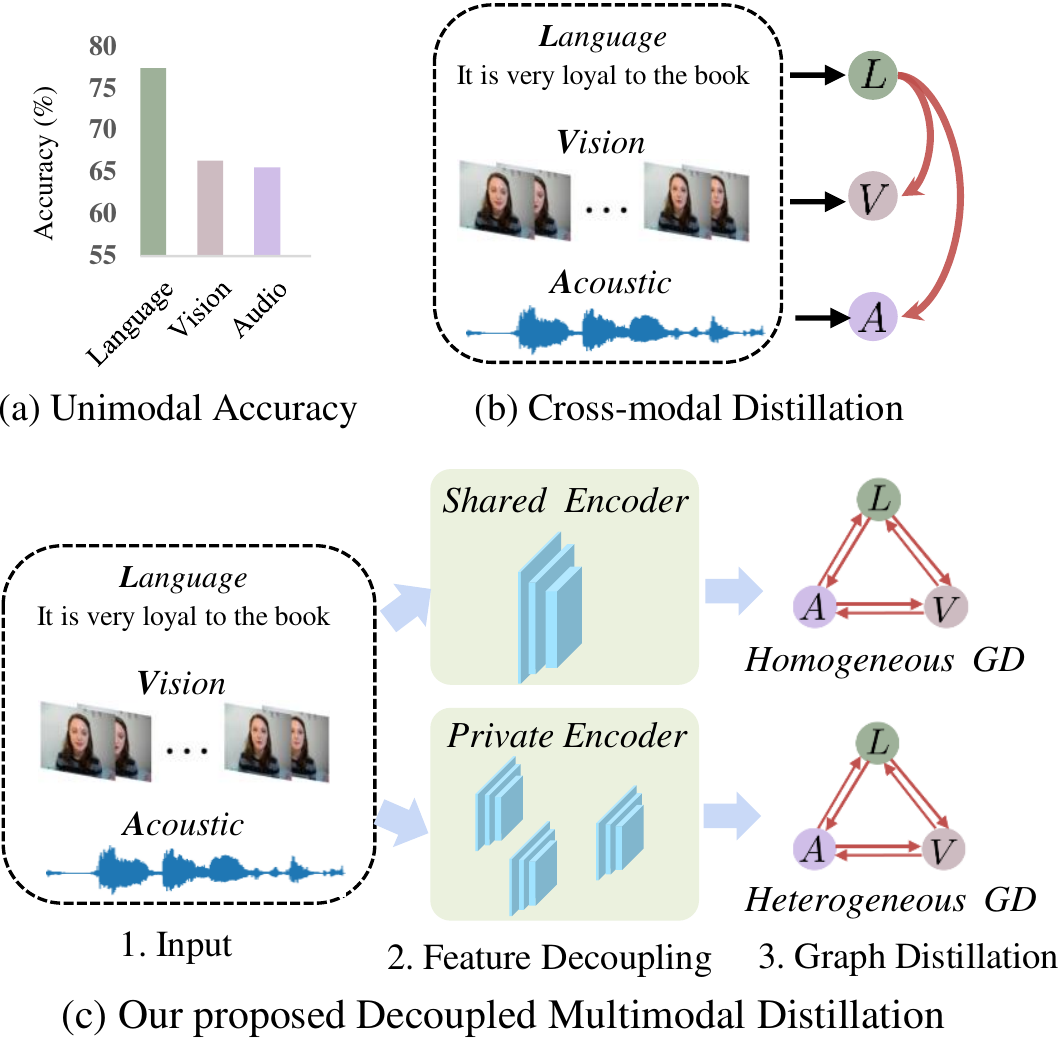}
	\caption{
		 (a) illustrates the significant emotion recognition discrepancies using unimodality, adapted from Mult~\cite{MulT}. (b) shows the conventional cross-modal distillation. (c) shows our proposed decoupled multimodal distillation (DMD) method. DMD consists of two graph distillation (GD) units: homogeneous GD and heterogeneous GD. The decoupled GD paradigm decreases the burden of absorbing knowledge from the heterogeneous data and allows each GD to be performed in a more specialized and effective manner.
	}
	\label{fig:main_idea}
\end{figure}

\footnote {*~The corresponding author.}

\section{Introduction}
\label{sec:intro}

Human multimodal emotion recognition (MER) aims to perceive the sentiment attitude of humans from video clips~\cite{PMR,MICA}. The video flows involve time-series data from various modalities, e.g., language, acoustic, and vision. This rich multimodality facilitates us in understanding human behaviors and intents from a collaborative perspective.  Recently, MER has become one of the most active research topics of affective computing with abundant appealing applications, such as intelligent tutoring systems~\cite{petrovica2017emotion}, product feedback estimation~\cite{melville2009sentiment}, and robotics~\cite{liu2017facial}.

For MER, different modalities in the same video segment are often complementary to each other, providing extra cues for semantic and emotional disambiguation. The core part of MER is multimodal representation learning and fusion, in which a model aims to encode and integrate representations from multiple modalities to understand the emotion behind the raw data. Despite the achievement of the mainstream MER methods~\cite{TFN,MulT,MISA}, the intrinsic heterogeneities among different modalities still perplex us and increase the difficulty of robust multimodal representation learning. Different modalities, e.g., image, language, and acoustic, contain different ways of conveying semantic information. Typically, the language modality consists of limited transcribed texts and has more abstract semantics than nonverbal behaviors.  As illustrated in Fig.~\ref{fig:main_idea} (a), language plays the most important role in MER and the intrinsic heterogeneities result in significant performance discrepancies among different modalities~\cite{MulT, pham2019found, TCSP}.


One way to mitigate the conspicuous modality heterogeneities is to distill the reliable and generalizable knowledge from the strong modality to the weak modality~\cite{Crossmodaldistillation}, as illustrated in Fig.~\ref{fig:main_idea} (b). However, such manual assignment for the distillation direction or weights should be cumbersome because there are various potential combinations. Instead, the model should learn to automatically adapt the distillation according to different examples, e.g, many emotions are easier to recognize via language while some are easier by vision. Furthermore, the significant feature distribution mismatch cross the modalities makes the direct crossmodal distillation sub-optimal~\cite{zhang2021matching, nguyen2021knowledge}.

To this end, we propose a decoupled multimodal distillation (DMD) method to learn dynamic distillations across modalities, as illustrated in Fig.~\ref{fig:main_idea} (c).  Typically, the features of each modality are decoupled into modality-irrelevant/-exclusive spaces via shared encoder and the private encodes, respectively. As to achieve the feature decoupling, we devise a self-regression mechanism that predicts the decoupled modality features and then regresses them self-supervisedly. To consolidate the feature decoupling, we incorporate a margin loss that regularizes the proximity in relationships of the representations across modalities and emotions. Consequently, the decoupled GD paradigm would decrease the burden of absorbing knowledge from the heterogeneous data and allows each GD to be performed in a more specialized and effective manner.

Based on the decoupled multimodal feature spaces, DMD utilizes a graph distillation unit (GD-Unit) in each space so that the crossmodal knowledge distillation can be performed in a more specialized and effective manner.
A GD-Unit consists of a graph that (1) vertices representing the representations or logits from the modalities and (2) edges indicating the knowledge distillation directions and weights.
As the distribution gap among the modality-irrelevant (homogeneous) features is sufficiently reduced,  GD can be directly applied to capture the inter-modality semantic correlations.
For the modality-exclusive (heterogeneous) counterparts, we exploit the multimodal transformer~\cite{MulT} to build the semantic alignment and bridge the distribution gap. The cross-modal attention mechanism in the multimodal transformer reinforces the multimodal representations and reduces the discrepancy between the high-level semantic concepts that exist in different modalities.
For simplification, we respectively name the distillation on the decoupled multimodal features as homogeneous graph knowledge distillation (HomoGD) and heterogeneous graph knowledge distillation (HeteroGD).  The reformulation allows us to explicitly explore the interaction between different modalities in each decoupled space.

The contributions of this work can be summarized as:
\begin{itemize}
	\item We propose a decoupled multimodal distillation framework, Decoupled Multimodal Distillation (DMD), to learn the dynamic distillations across modalities for robust MER. 
	In DMD, we explicitly decouple the multimodal representations into modality-irrelevant/-exclusive spaces to facilitate KD on the two decoupled spaces.
	DMD provides a flexible knowledge transfer manner where the distillation directions and weights can be automatically learned, thus enabling flexible knowledge transfer patterns.
	\item We conduct comprehensive experiments on public MER datasets and obtain superior or comparable results than the state-of-the-arts. Visualization results verify the feasibility of DMD and the graph edges exhibit meaningful distributional patterns w.r.t. HomoGD and  HeteroGD.
\end{itemize}

\section{Related Works}
\label{sec:RL}

\subsection{Multimodal emotion recognition}
Multimodal emotion recognition (MER) aims to infer human sentiment from the language, visual and acoustic information embedded in the video clips.
The heterogeneity across modalities can provide various levels of information for MER.
The mainstream MER approaches can be divided into two categories: fusion strategy-based \cite{TFN,LMF,MFN} and crossmodal attention-based \cite{MulT,PMR,MICA}.

The former aims to design sophisticated multimodal fusion strategies to generate discriminative multimodal representations, e.g., Zadeh \emph{et al.} \cite{TFN} designed a Tensor Fusion Network (TFN) that can fuse multimodal information progressively.
However, the inherent heterogeneity and the intrinsic information redundancy across modalities hinders the fusion between the multimodal features.
Therefore, some work aims to explore the characteristics and commonalities of multimodal representations via feature decoupling to facilitate more effective multimodal representation fusion \cite{MFM,MISA,FDMER}.
Hazarika \emph{et al.} \cite{MISA} decomposed multimodal features into modality-invariant/-specific components to learn the refined multimodal representations. The decoupled multimodal representations reduce the information redundancy and provide a holistic view of the multimodal data.
Recently, crossmodal attention-based approaches have driven the development of MER  since they learn the cross-modal correlations to obtain the reinforced modality representation. A representative work is MulT~\cite{MulT}. This work proposes the multimodal transformer that consists of a cross-modal attention mechanism to learn the potential adaption and correlations from one modality to another, thereby achieving semantic alignment between modalities.
Lv \emph{et al.} \cite{PMR} designed a progressive modality reinforcement method based on \cite{MulT}, it aims to learn the potential adaption and correlations from multimodal representation to unimodal representation. Our proposed DMD has an essential difference with the previous feature-decoupling methods~\cite{MFM,MISA,FDMER} because DMD is capable of distilling cross-modal knowledge in the decoupled feature spaces.

\begin{figure*}[htb]
	\centering{\includegraphics[width=0.95\linewidth]{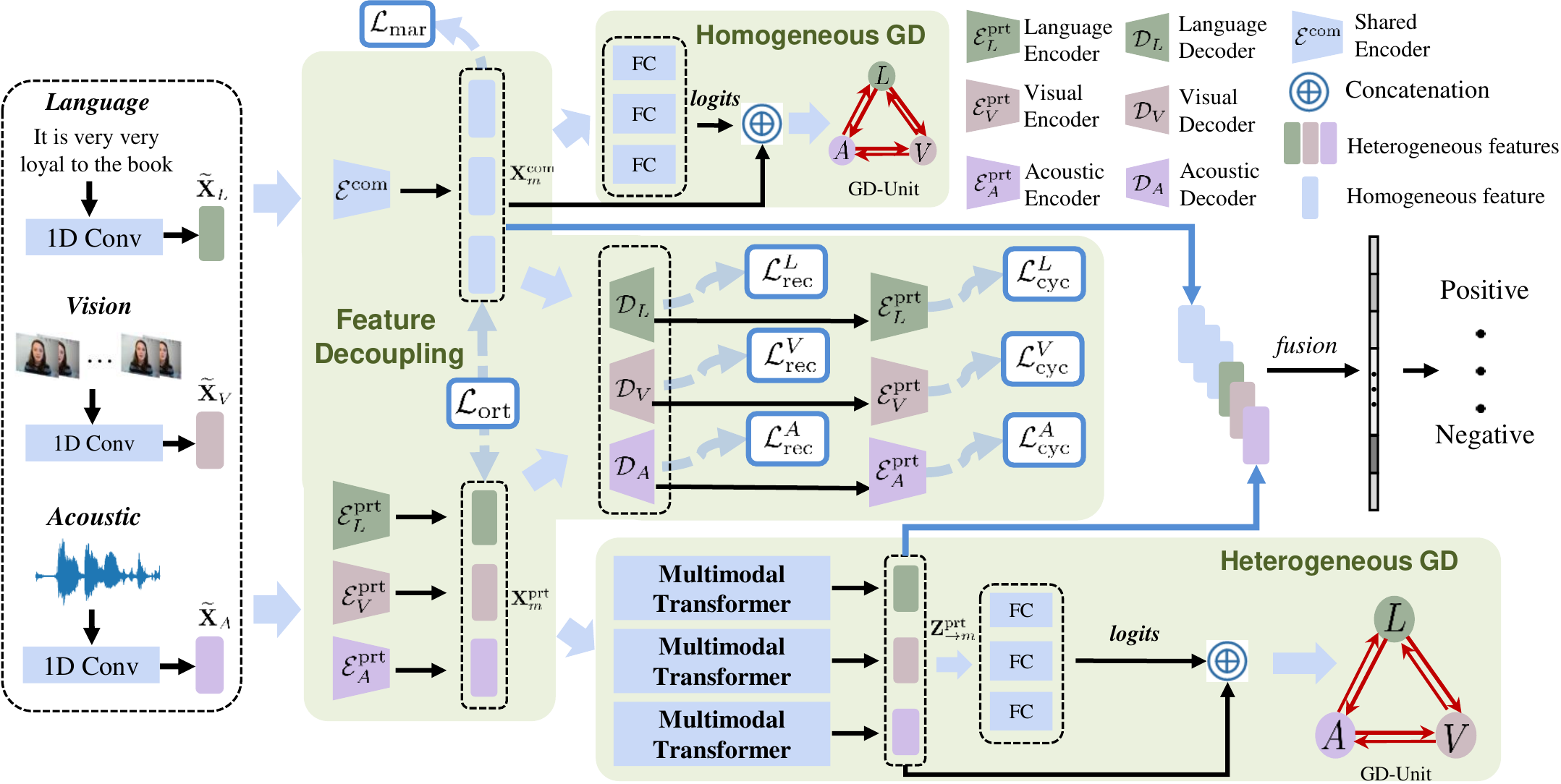}}
	\caption{The framework of DMD. Given the input multimodal data, DMD encodes their respective shallow features $\tbX_{m}$, where $m \in \{L, V, A\}$.
		In \textit{feature decoupling}, DMD exploits the decoupled homo-/heterogeneous multimodal features 
		$\X^{\text{com}}_{m}$ / $\X^{\text{prt}}_{m}$ via the shared and exclusive encoders, respectively. $\X^{\text{prt}}_{m}$ will be reconstructed in a self-regression manner (Sec.~\ref{sec:decoupling}). 
		Subsequently, $\X^{\text{com}}_{m}$ will be fed into a GD-Unit for adaptive knowledge distillation in \textit{HomoGD}.
		In \textit{HeteroGD}, $\X^{\text{prt}}_{m}$ are reinforced to $\Z^{\text{prt}}_{\to m}$ via multimodal transformers to bridge the distribution gap. The GD-Unit in \textit{HeteroGD} takes $\Z^{\text{prt}}_{\to m}$ as input for distillation (Sec.~\ref{sec:distillation}). Finally, $\X^{\text{com}}_{m}$ and $\Z^{\text{prt}}_{\to m}$ will be adaptively fused for MER. 
	}
	\label{fig:framework}
	\vspace{-10pt}
\end{figure*}

\subsection{Knowledge distillation}
The concept of knowledge distillation (KD) was first proposed in \cite{KD} to transfer knowledge via minimizing the KL-Divergence between prediction logits of teachers and students.
Subsequently, various KD methods were proposed \cite{logitsKD1,logitsKD2,logitsKD3,logitsKD4} based on \cite{KD} and further extended to distillation between intermediate features \cite{featKD1,featKD2,featKD3,featKD4}.

Most KD methods focus on transferring knowledge from the teacher to the student, while some recent studies have used graph structures to explore the effective message passing mechanism between multiple teachers and students with multiple instances of knowledge \cite{GDIJCAI,GDECCV,GDACCV}.
Zhang \textit{et al.} \cite{GDIJCAI} proposed a graph distillation (GD) method for video classification, where each vertex represented a self-supervised teacher and edges represented the direction of distillation from multiple self-supervised teachers to the student.
Luo \emph{et al.} \cite{GDECCV} considered the modality discrepancy to incorporate privileged information from the source domain and modeled a directed graph to explore the relationship between different modalities. Each vertex represented a modality and the edges indicated the connection strength (i.e., distillation strength) between one modality and another.
Different from them, we aim to use exclusive GD-Units in the decoupled feature spaces to facilitate effective cross-modality distillation.

\section{The Proposed Method}

The framework of our DMD is illustrated in Fig.~\ref{fig:framework}.
It mainly consists of three parts: \textbf{multimodal feature decoupling}, \textbf{homogeneous GD} (HomoGD), \textbf{hetergeneous GD} (HeteroGD). Considering the significant distribution mismatch of modalities, we decouple multimodal representations into homogeneous and heterogeneous multimodal features through learning shared and exclusive multimodal encoders. The decoupling detail is introduced in Sec.~\ref{sec:decoupling}. To facilitate a flexible knowledge transfer, we next distill the knowledge from homo/heterogeneous features, which are framed in two graph distillation units (GD-Unit), i.e., HomoGD and HeteroGD.
In HomoGD, homogeneous multimodal features are mutually distilled to compensate the representation ability for each other. In HeteroGD, multimodal transformers are introduced to explicitly build inter-modal correlations and semantic alignment for further distilling. The GD detail is introduced in Sec.~\ref{sec:distillation}.  Finally, the refined multimodal features through distilling are adaptively fused for robust MER.
Below, we present the details of the three parts of DMD.

\subsection{Multimodal feature decoupling}
\label{sec:decoupling}

We consider three modalities, i.e., \textit{language} (L), \textit{visual} (V), \textit{acoustic} (A).
Firstly, we exploit three separate 1D temporal convolutional layers to aggregate temporal information and obtain the low-level multimodal features: $\tbX_{m} \in \mathbb{R}^{T_{m} \times d_m}$, where $m \in \{L, V, A\}$ indicates a modality. After this shallow encoding, each modality preserves the input temporal dimension to facilitate handling unaligned and aligned cases simultaneously. Moreover, all modalities are scaled to the same feature dimension, i.e., $d_L=d_V=d_A=d$, for convenient subsequent feature decoupling.

To decouple the multimodal features into homogeneous (modality-irrelevant) part $\X_{m}^\sha$ and heterogeneous (modality-exclusive) part $\X_{m}^\prt$, we exploit a shared multimodal encoder $\mathcal{E}^\sha$ and three private encoders $\mathcal{E}_{m}^\prt$ to explicitly predict the decoupled features. Formally,
\begin{equation}
	\X_{m}^\sha = \mathcal{E}^\sha(\tbX_{m}), \X_{m}^\prt=\mathcal{E}_{m}^\prt(\tbX_{m}).
\end{equation}
To distinguish the differences between $\X_{m}^\sha$ and $\X_{m}^\prt$ and mitigate the feature ambiguity, we synthesize the vanilla coupled features $\tbX_m$ in a self-regression manner. Mathematically speaking, we concatenate $\X_{m}^\sha$ and $\X_{m}^\prt$ for each modality and exploit a private decoder $\mathcal{D}_{m}$ to produce the coupled feature, i.e., $\mathcal{D}_{m}([\X_{m}^\sha, \X_{m}^\prt])$. Subsequently, the coupled feature  will be re-encoded via the private encoders $\mathcal{E}_{m}^\prt$ to regress the heterogeneous features. The notation $[.]$ means feature concatenation. Formally, the discrepancy between the vanilla/synthesized coupled multimodal features can be formulated as:
\begin{align}
	\mathcal{L}_{\text{rec}} = \| \tbX_m - \mathcal{D}_{m}([\X_{m}^\sha, \X_{m}^\prt]) \|_F^2.\label{equ:modality_rec}
\end{align}
Further, the  discrepancy between the vanilla/synthesized heterogeneous features can be formulated as:
\begin{align}
	\mathcal{L}_{\text{cyc}} = \| \X_{m}^\prt - \mathcal{E}_{m}^\prt(\mathcal{D}_{m}([\X_{m}^\sha, \X_{m}^\prt]))\|_F^2.
\end{align}

For the above reconstruction losses, it still cannot guarantee the complete feature decoupling. In fact, information can freely leak between representations, e.g., all the modality information can be merely encoded in $\X_{m}^\prt$ so that the decoders can easily synthesize the input, leaving homogeneous multimodal features meaningless.
To consolidate the feature decoupling, we argue that homogeneous representations from the same emotion but different modalities should be more similar than those from the same modality but different emotions. To this end, we define a margin loss as:
	\begin{align}
		&\!\!\!\mathcal{L}_{\text{mar}} =\nonumber\\
  &\!\!\!\!\!\!\frac{1}{|S|}\!\!\!\sum_{(i,j,k)\in S}\!\!\!\!\!\max(0, \alpha\!\! -\!\! \cos( {\X_{m[i]}^\sha, \X_{m[j]}^\sha} )\!\!+\!\!\cos( {\X_{m[i]}^\sha, \X_{m[k]}^\sha} ) ),
		\label{equ:homo_constraint}
	\end{align}
where we collect a triple tuple set $S=\{(i,j,k)| m[i]\neq m[j], m[i]= m[k], c[i]= c[j], c[i]\neq c[k] \}$. The $m[i]$ is the modality of sample $i$, $c[i]$ is the class label of sample $i$, and $\cos(\cdot,\cdot)$ means the cosine similarity between two feature vectors. The loss in Eq.~\ref{equ:homo_constraint} constrains the homogeneous features that belong to the same emotion but different modalities or vice versa to differ, and thereby avoids deriving trivial homogeneous features. 
$\alpha$ is a distance margin. The distances of positive samples (same \textit{emotion}; different \textit{modalities})  are constrained to be smaller than that of negative samples (same \textit{modality}; different \textit{emotions}) by the margin $\alpha$.
Considering that the decoupled features respectively capture the modality-irrelevant/-exclusive characteristics, we further formulate a soft orthogonality to reduce the information redundancy between the homogeneous and the heterogeneous multimodal features:
\begin{equation}
	\mathcal{L}_{\text{ort}} = {\sum\limits_{m \in \{ L,V,A\}}{\cos(\X_{m}^\sha, \X_{m}^\prt)}}.
	\label{equ:orthogonality}
\end{equation}
Finally, we combine these constraints to form the decoupling loss,
\begin{align}
\mathcal{L}_{\text{dec}} = \mathcal{L}_{\text{rec}} + \mathcal{L}_{\text{cyc}} + \gamma(\mathcal{L}_{\text{mar}} + \mathcal{L}_{\text{ort}}),
\end{align}
where $\gamma$ is the balance factor.


\subsection{GD with Decoupled Multimodal Features}
\label{sec:distillation}

For the decoupled homogeneous and heterogeneous multimodal features, we design a graph distillation unit (GD-Unit) on each of them to conduct adaptive knowledge distillation. Typically, a GD-Unit consists of a directed graph $\mathcal{G}$. Let $v_i$ denote a node w.r.t a modality and $w_{i \rightarrow j}$ indicates the distillation strength from modality $v_i$ to $v_j$. The distillation from $v_i$ to $v_j$ is defined as the difference between their corresponding logits, denoted with $\epsilon_{i \rightarrow j}$. Let $\E$ denotes the distillation matrix with  $E_{ij} = \epsilon_{i \rightarrow j}$. For a target modality $j$, the weighted distillation loss can be formulated by considering the injection edges as,
\begin{equation}
	\zeta_{:j} = \sum_{v_i \in \mathcal{N}(v_j)} w_{i \rightarrow j} \times \epsilon_{i \rightarrow j},
\end{equation}
where $\mathcal{N}(v_j)$ represents the set of vertices injected to $v_j$.

To learn a dynamic and adaptive weight that corresponds to the distillation strength $w$, we propose to encode the modality logits and the representations into the graph edges. Formally,
\begin{align}
w_{i \rightarrow j}= g([[{f(\X_{i},\theta_1)}, {\X}_{i}], [{f(\X_{j},\theta_1)}, {\X_{j}}]], \theta_2),
	\label{equ:dynamic_edge}
\end{align}
where $[\cdot,\cdot]$ means feature concatenation, $g$ is a fully-connected (FC) layer with the learnable parameters $\theta_2$, and $f$ is a FC layer for regressing logits with the parameters $\theta_1$.   
The graph edge weights $\W$ with $W_{ij}=w_{i\rightarrow j}$ can be constructed and learned by repetitively applying Eq.~\ref{equ:dynamic_edge} over all pairs of modalities. To reduce the scale effects, we normalize $\W$ through the $softmax$ operation. Consequently, the graph distillation loss to all modalities can be written as:
\begin{align}
	\mathcal{L}_{\text{dtl}} = \|\W \odot \E\|_1,
	\label{equ:overall_distillation_loss}
\end{align}
where $\odot$ means element-wise product. Obviously, the distillation graph in a GD-Unit provides a base for learning dynamic inter-modality interactions. Meanwhile, it facilitates a flexible knowledge transfer manner where the distillation strengths can be automatically learned, thus enabling diverse knowledge transfer patterns.
Below, we elaborate on the details of HomoGD and HeteroGD.

\textbf{HomoGD.}
As illustrated in Fig.~\ref{fig:framework}, for the decoupled homogeneous features $\X_m^\sha$, as the distribution gap among the modalities is already reduced sufficiently, we input the features $\X_m^\sha$ and the corresponding logits $f(\X_m^\sha)$ to a GD-Unit and calculate the graph edge matrix $\W$ and the distillation loss matrix $\E$ according to Eq.~\ref{equ:dynamic_edge}. Then, the overall distillation loss $\mathcal{L}_{\text{dtl}}^{\text{homo}}$ can be naturally obtained via Eq.~\ref{equ:overall_distillation_loss}.

\textbf{HeteroGD.}
The decoupled heterogeneous features $\X_m^\prt$ focus on the diversity and the unique characteristics of each modality, and thus exhibit a significant distribution gap.
To mitigate this issue, we exploit the multimodal transformer~\cite{MulT} to bridge the feature distribution gap and build the modality adaptation.
The core of the multimodal transformer is the crossmodal attention unit ($\text{CA}$), which receives features from a pair of modalities and fuses crossmodal information. 
Take the language modality $\X_L^\prt$ as the source and the visual modality $\X_V^\prt$ as the target, the cross-modal attention can be defined as: $\Q_{V}=\X_V^\prt \P_{q}$, $\K_{L}=\X_L^\prt \P_{k}$, and $\V_{L}=\X_L^\prt \P_{v}$ where $\P_q, \P_k, \P_v$ are the learnable parameters. The individual head of can be expressed as:
	\begin{align}
		\Z_{L \to V}^\prt&= \text{softmax}(\frac{\Q_{V}\K_{L}^{\top}}{\sqrt{d}})\V_L,
	\end{align}
where $\Z_{L \to V}^\prt$ is the enhanced features from Language to Visual, $d$ means the dimension of $\Q_{V}$ and $\K_{L}$. For the three modalities in MER, each modality will be reinforced by the two others and the resulting features will be concatenated. For each target modality, we concatenate all enhanced features from other modalities to the target as the reinforced features, denotes with $\Z^\prt_{\rightarrow m}$, which are used in the distillation loss function as $\mathcal{L}_{\text{dtl}}^{\text{hetero}}$ that can be naturally obtained via Eq.~\ref{equ:overall_distillation_loss}.

\textbf{Feature fusion.} We use the reinforced heterogeneous features $\Z^\prt_{\rightarrow m}$ and the vanilla decoupled homogeneous features $\X_m^\sha$ for adaptive feature fusion with an adaptive weight learned from each of them. Hereby, we obtain the fused feature for multimodal emotion recognition.

\subsection{Objective optimization}

We integrate the above losses to reach the full objective:
\begin{align}
	\mathcal{L}_{\text{total}} = \mathcal{L}_{\text{task}} + \lambda_1 \mathcal{L}_{\text{dec}} + \lambda_2\mathcal{L}_{{\text{dtl}}},
\end{align}
where $\mathcal{L}_{\text{task}}$ is the emotion task related loss (here mean absolute error),
$\mathcal{L}_{\text{dtl}} = \mathcal{L}_{\text{dtl}}^{\text{homo}} + \mathcal{L}_{\text{dtl}}^{\text{hetero}}$ means the distillation losses generated by HomoGD and HeteroGD, and $\lambda_1, \lambda_2$ control the importance of different constraints.

				

\section{Experiments}

\textbf{Datasets.}
We evaluate DMD on CMU-MOSI \cite{mosi} and CMU-MOSEI \cite{mosei} datasets. The experiments are conducted under the word-aligned  and unaligned settings for a more comprehensive comparison.
\textbf{CMU-MOSI} consists of 2,199 short monologue video clips. 
The acoustic and visual features in CMU-MOSI are extracted at a sampling rate of 12.5 and 15 Hz, respectively.
Among the samples, 1,284, 229 and 686 samples are used as training, validation and testing set.
\textbf{CMU-MOSEI} contains 22,856 samples of movie review video clips from YouTube (approximately $10 \times$ the size of CMU-MOSI).
The acoustic and visual features were extracted at a sampling rate of 20 and 15 Hz, respectively.
According to the predetermined protocol,  16,326 samples are used for training, the remainng 1,871 and 4,659 samples are used for for validation and testing.
Each sample in CMU-MOSI and CMU-MOSEI was labeled with a sentiment score which ranges from -3 to 3, including \textit{highly negative}, \textit{negative}, \textit{weakly negative}, \textit{neutral}, \textit{weakly positive}, \textit{positive}, and \textit{highly positive}.
Following previous work \cite{PMR, MICA}, we evaluate the MER performance using the following metrics: 7-class accuracy (ACC$_{7}$), binary accuracy (ACC$_{2}$) and F1 score.

\textbf{Implementation details.} On the two datasets, we extract the unimodal language features via GloVe \cite{GloVe} and obtain 300-dimensional word features.
For a fair comparison with ~\cite{MISA, FDMER} under the aligned setting, we additationally exploit a BERT-base-uncased pre-trained model~\cite{bert} to obtain a 768-dimensional hidden state as the word features.
For visual modality, each video frame was encoded via Facet~\cite{Facet} to represent the presence of the totally 35 facial action units~\cite{li2019self, li2020learning}. 
The acoustic modality was processed by COVAREP~\cite{COVAREP} to obtain the 74-dimensional features.
\textit{The detailed neural network configurations in DMD are listed in the supplementary file.}
The optimal setting for $\lambda_1$, $\lambda_2$, $\gamma$ was set as 0.1, 0.05, 0.1 via the MER performance on the validation set.
We implemented all the experiments using PyTorch on a RTX 3090 GPU with 24GB memory. We set the training batch size as 16 and trained DMD for 30 epoches until convergence.

\begin{table}[t]
	\centering
	\setlength{\tabcolsep}{2pt}
	\caption{Comparison on CMU-MOSI dataset. \textbf{Bold} is the best.}\label{tab:MOSI}
	\vspace{-0.4cm}
	\scalebox{0.85}{
		\begin{tabular}{c|c|ccc}
			\hline
			Methods & Setting  &  ACC$_{7}$ (\%) & ACC$_{2}$ (\%) & F1 (\%)   \\
			\cline{3-5}
			\hline
			\hline
			EF-LSTM& \multirow{12}{*}{Aligned} &   33.7&  75.3&  75.2\\
			LF-LSTM &&  35.3& 76.8&76.7  \\
			TFN \cite{TFN} & & 32.1& 73.9& 73.4\\
			LMF \cite{LMF} & & 32.8& 76.4& 75.7\\
			MFM \cite{MFM} & & 36.2& 78.1& 78.1\\
			RAVEN \cite{RAVEN} && 33.2& 78.0 & 76.6\\
			MCTN \cite{MCTN} && 35.6& 79.3& 79.1\\
			MulT \cite{MulT} & & 40.0&  83.0& 82.8\\
			PMR \cite{PMR} & & 40.6 & 83.6& 83.4\\
			
			DMD (\textbf{Ours}) && \textbf{41.4} & \textbf{84.5} & \textbf{84.4} \\
			\hline
			MISA \cite{MISA}$^{*}$ & \multirow{3}{*}{Aligned}  & 42.3& 83.4& 83.6\\
			FDMER \cite{FDMER}$^{*}$ & & 44.1& 84.6& 84.7\\
			DMD (\textbf{Ours})$^{*}$ &&\textbf{45.6} &\textbf{86.0}  &\textbf{86.0} \\
			\hline
			\hline
			EF-LSTM& \multirow{8}{*}{Unaligned} &  31.0&  73.6&  74.5\\
			LF-LSTM && 33.7 & 77.6&77.8   \\
			RAVEN \cite{RAVEN} & &31.7    & 72.7 & 73.1\\
			MCTN \cite{MCTN} & & 32.7  & 75.9& 76.4\\
			MulT \cite{MulT} & &  39.1 & 81.1& 81.0\\
			PMR \cite{PMR} & & 40.6  & 82.4& 82.1\\
			MICA \cite{MICA}& &  40.8& 82.6& 82.7\\
			DMD (\textbf{Ours}) && \textbf{41.9}&\textbf{83.5}  &\textbf{83.5}\\
			\hline
	\end{tabular}}
	\begin{tablenotes}
		\centering
		\footnotesize
		\item[1] * means the input language features are BERT-based.
	\end{tablenotes}
\end{table}

\begin{table}[t]
	\centering
	\setlength{\tabcolsep}{2.5pt}
	\caption{Comparison on CMU-MOSEI dataset. \textbf{Bold} is the best. }\label{tab:MOSEI}
	\vspace{-0.4cm}
	\scalebox{0.85}{
	\begin{tabular}{c|c|ccc}
		\hline
		Methods & Setting  &  ACC$_{7}$ (\%) & ACC$_{2}$ (\%) & F1 (\%)   \\
		\cline{3-5}
		\hline
		\hline
		EF-LSTM& \multirow{8}{*}{Aligned} &  47.4 &  78.2& 77.9 \\
		LF-LSTM &&  48.8& 80.6&   80.6 \\
		Graph-MFN \cite{mosei} & & 45.0& 76.9&  77.0\\
		RAVEN \cite{RAVEN} && 50.0&  79.1&  79.5 \\
		MCTN \cite{MCTN} && 49.6& 79.8& 80.6 \\
		MulT \cite{MulT} & & 51.8& 82.5 & 82.3\\
		PMR \cite{PMR} & & 52.5& 83.3& 82.6\\
		DMD (\textbf{Ours}) &&\textbf{53.7}&\textbf{85.0}  &\textbf{84.9} \\
		
		\hline
		MISA \cite{MISA}$^{*}$ & \multirow{3}{*}{Aligned}  &  52.2&  85.5& 85.3 \\
		FDMER \cite{FDMER}$^{*}$ & &  54.1& 86.1& 85.8 \\
		DMD (\textbf{Ours})$^{*}$ & &\textbf{54.5} &\textbf{86.6}  &\textbf{86.6} \\
		
		\hline
		\hline
		EF-LSTM& \multirow{8}{*}{Unaligned} &  46.3   &  76.1&75.9  \\
		LF-LSTM &&  48.8  & 77.5&  78.2 \\
		RAVEN \cite{RAVEN} & &45.5    &  75.4&75.7 \\
		MCTN \cite{MCTN} & & 48.2  & 79.3& 79.7\\
		MulT \cite{MulT} & & 50.7  & 81.6& 81.6\\
		PMR \cite{PMR} & & 51.8  & 83.1& 82.8\\
		MICA \cite{MICA}& & 52.4  & 83.7& 83.3\\
		DMD (\textbf{Ours}) &&\textbf{54.6}&\textbf{84.8}  &\textbf{84.7}\\
		\hline
	\end{tabular}}
	\begin{tablenotes}
		\centering
		\footnotesize
		\item[1] * means the input language features are BERT-based.
	\end{tablenotes}
\end{table}

\begin{table}[t]
	\centering
	\setlength{\tabcolsep}{2.5pt}
	\caption{Ablation study of the key components in DMD.}\label{tab:Ablation}
	\vspace{-0.4cm}
	\scalebox{0.85}{
		\begin{tabular}{c|cccc|cc}
			\hline
			Dataset & FD & HomoGD &  CA & HeteroGD  & ACC$_{7}$ & F1 \\
			\hline
			\hline
			\multirow{6}{*}{MOSI} & $\checkmark$ & $\checkmark$ & $\checkmark$ &  $\checkmark$ & \textbf{41.9}& \textbf{83.5}\\
			& $\checkmark$ & $\checkmark$ & $\checkmark$ &  $\times$ & 38.8 & 81.1 \\
			&  $\checkmark$ & $\checkmark$ & $\times$ &  $\checkmark$ & 37.5& 80.6\\
			&  $\checkmark$ & $\checkmark$ & $\times$ &  $\times$ & 37.2& 80.8\\
			&  $\checkmark$ & $\times$ & $\times$ &  $\times$ & 34.7& 79.3\\
			&  $\times$ & $\times$ & $\times$ &  $\times$ & 32.4& 79.0\\
			\hline
			\hline
			\multirow{6}{*}{MOSEI}  & $\checkmark$ & $\checkmark$ & $\checkmark$ &  $\checkmark$ & \textbf{54.6}& \textbf{84.7}\\
			& $\checkmark$ & $\checkmark$ & $\checkmark$ &  $\times$ & 53.2& 84.1\\
			&  $\checkmark$ & $\checkmark$ & $\times$ &  $\checkmark$ & 52.4&  83.8\\
			&  $\checkmark$ & $\checkmark$ & $\times$ &  $\times$ & 52.4& 84.3\\
			&  $\checkmark$ & $\times$ & $\times$ &  $\times$ & 51.6&  82.8\\
			&  $\times$ & $\times$ & $\times$ &  $\times$ & 50.0& 81.9\\
			\hline
	\end{tabular}}
\end{table}

\begin{table}[htb]
	\begin{center}
		\caption{Unimodal accuracy comparison on MOSEI dataset.}\label{tab:fd}
		\vspace{-0.4cm}
		\scalebox{0.75}{
			\begin{tabular}{c|c|c}
				\hline
				\multirow{2}{*}{Methods} &
				w/o FD & w/ FD\\
				\cline{2-3}
				& Acc$_{2}$ (\%)~/~F1 (\%) & Acc$_{2}$ (\%)~/~F1 (\%)   \\
				\hline
				\hline
				$L$ only &81.2~~/~~81.4 & \textbf{82.7~~/~~82.5}
				\\
				$V$  only & 58.2~~/~~52.2& \textbf{62.8~~/~~60.0}
				\\
				$A$  only & 53.4~~/~~54.0 & \textbf{64.9~~/~~62.5}
				\\
				\hline
				Mean & 64.3~~/~~62.5& \textbf{70.1~~/~~68.3}
				\\
				STD & 12.1~~/~~13.4& \textbf{8.9~~/~~10.1}
				\\
				\hline
		\end{tabular}}
	\end{center}
\end{table}

\begin{table}[t]
	\centering
	\caption{Ablation study of graph distillation (GD) on MulT.}\label{tab:mult with gd}
	\vspace{-0.4cm}
	\setlength{\tabcolsep}{1.5pt}
	\scalebox{0.8}{
		\begin{tabular}{c|ccc|ccc}
			\hline
			\multirow{2}{*}{Methods} & \multicolumn{3}{c|}{CMU-MOSI} & \multicolumn{3}{c}{CMU-MOSEI}\\
			\cline{2-7} &
			ACC$_{7}$  & ACC$_{2}$ & F1 &
			ACC$_{7}$ & ACC$_{2}$ & F1   \\
			\hline
			\hline
			MulT & 39.1& 81.1 & 81.0 & 50.7 & 81.6 & 81.6 \\
			MulT (\textit{w/ GD}) & 39.4 & 82.2 & 82.2 & 51.0 & 82.3 & 82.5  \\
			DMD (\textbf{Ours}) & \textbf{41.9} & \textbf{83.5} & \textbf{83.5} & \textbf{54.6} & \textbf{84.8} & \textbf{84.7}  \\
			\hline
	\end{tabular}}
\end{table}

\begin{figure*}[t]
	\centering{\includegraphics[width=0.8\linewidth]{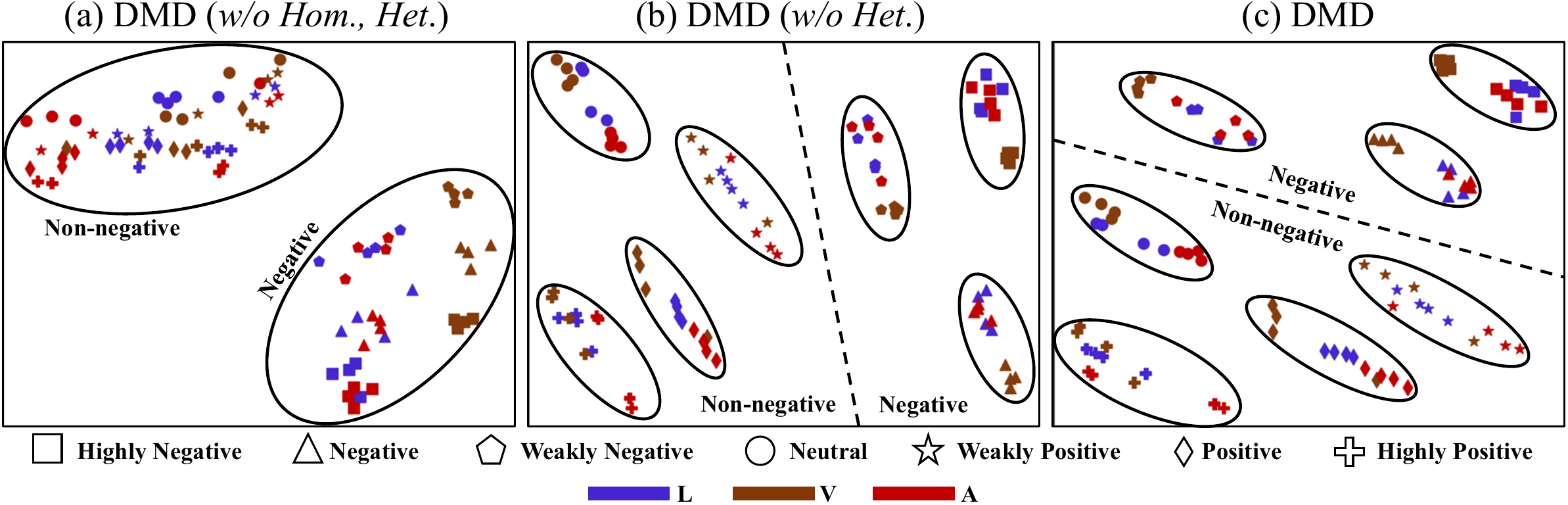}}
	\caption{t-SNE visualization of decoupled homogeneous space on MOSEI. DMD shows the promising emotion category (binary or 7-class)  separability in (c).	}
	\label{fig:tsne-c}
		\vspace{-5pt}
\end{figure*}

\begin{figure}[t]
	\centering{\includegraphics[width=\linewidth]{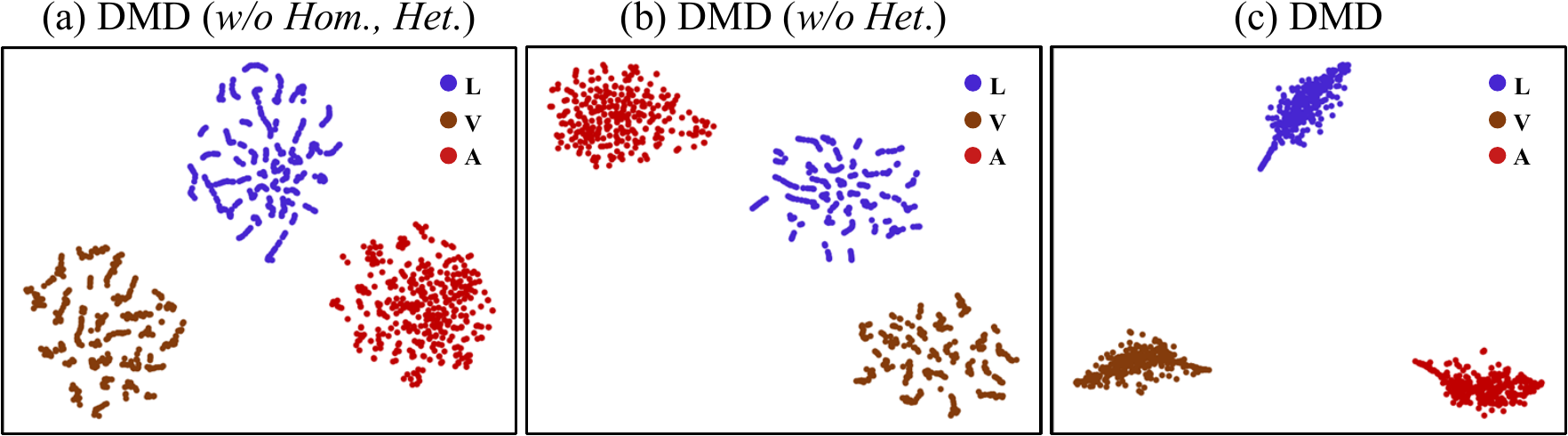}}
	\caption{Visualization of the decoupled heterogeneous features on MOSEI.
    DMD shows the best modality separability in (c).
	}
	\label{fig:tsne-s}
		\vspace{-5pt}
\end{figure}

\subsection{Comparison with the state-of-the-art}
We compare DMD with the current state-of-the-art MER methods under the same dataset settings (unaligned or aligned), including EF-LSTM, LF-LSTM, TFN~\cite{TFN}, LMF~\cite{LMF}, MFM~\cite{MFM}, RAVEN~\cite{RAVEN}, Graph-MFN~\cite{mosei}, MCTN~\cite{MCTN}, MulT~\cite{MulT}, PMR~\cite{PMR}, MICA~\cite{MICA}, MISA~\cite{MISA}, and FDMER~\cite{FDMER}.

Tab.~\ref{tab:MOSI} and Tab.~\ref{tab:MOSEI} illustrate the comparison on CMU-MOSI and CMU-MOSEI datasets, respectively.
Obviously, our proposed DMD obtains superior MER accuracy than other MER approaches under the unaligned and aligned settings.
Compared with the feature-disentangling-based MER methods~\cite{MISA, FDMER, MFM}, our proposed DMD obtains consistent improvements, indicating the feasibility of the incorporated GD-Unit, which is capable of perceiving the various inter-modality dynamics. For a further inverstigation, we will visualize the learned graph edges in each GD-Unit in Sec.~\ref{sec:ablation_study}.
DMD consistently outperforms the methods~\cite{MulT, MICA, PMR} that use multimodal transformer to learn crossmodal interactions and perform multimodal fusion. The reasons are two-fold: (1) DMD takes the modality-irrelevant/-exclusive spaces into consideration concurrently and recudes the information redundancy via feature decoupling. (2) DMD exploits twin GD-Units to adaptively distil knowledge among the modalities.
On CMU-MOSEI dataset, Graph-MFN~\cite{mosei} illustrates unsatisfactory results because the heterogeneity and distribution gap across modalities hinder the learning of the modality fusion in it. As a comparison, the multimodal features into DMD are decoupled into modality-irrelevant/-exclusive spaces.  For the latter space, we use multimodal transformer to bridge the distribution gap and align the high-level semantics, thereby reducing the burden of absorbing knowledge from the heterogeneous features.

\begin{figure*}[t]
	\centering{\includegraphics[width=0.85\linewidth]{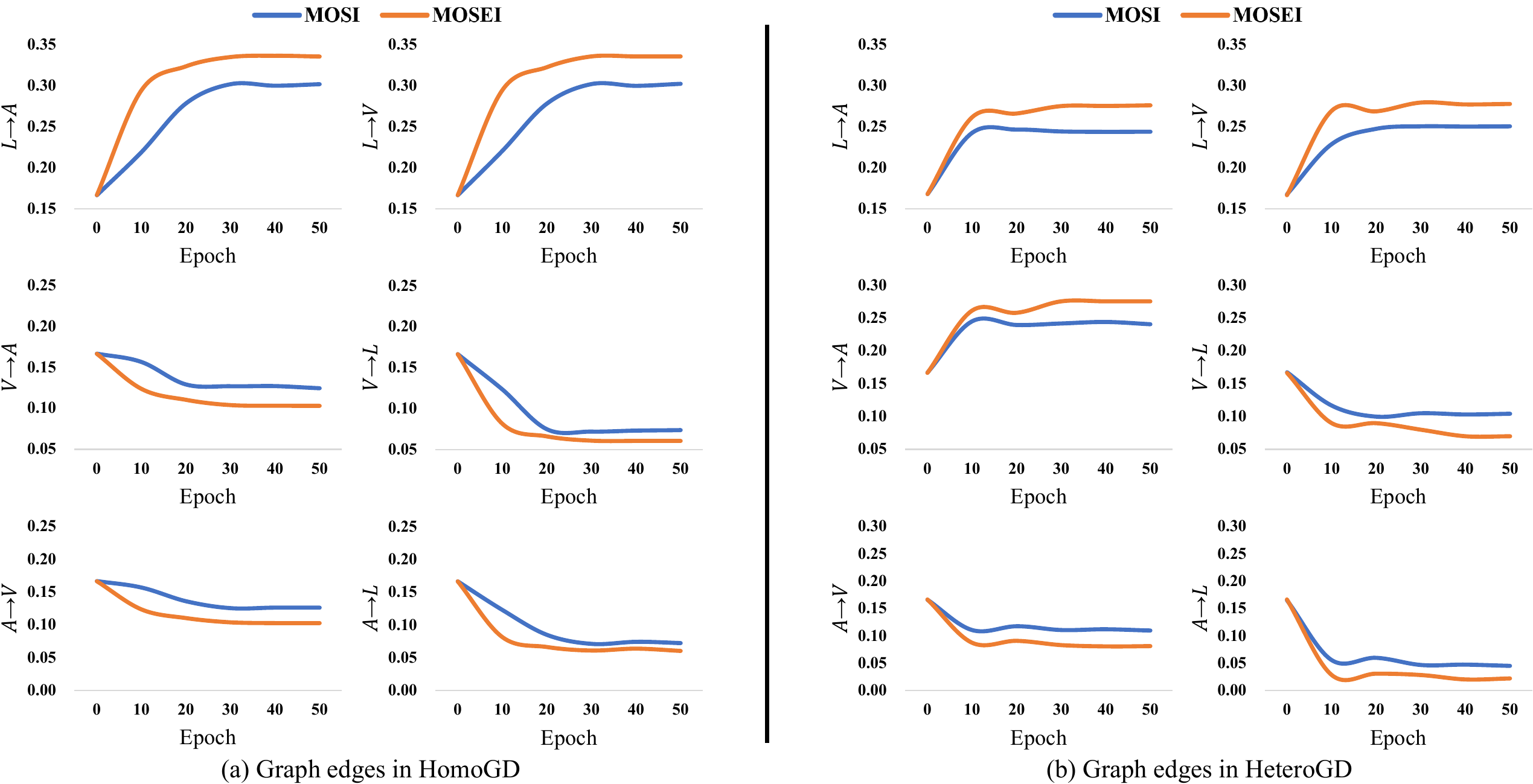}}
	\caption{Illustration of the graph edges in HomoGD and HeteroGD. In (a), $L \to A$ and $L \to V$ are dominated because the homogeneous language features contribute most and the other modalities perform poorly.
	In (b), $L \to A$, $L \to V$, and $V \to A$ are dominated.  $V \to A$ emerges because the \textit{visual} modality enhanced its feature discriminability via the multimodal transformer mechanism in HeteroGD.}
	\label{fig:edge}
\end{figure*}

\subsection{Ablation study}
\label{sec:ablation_study}
\textbf{Quantitative analysis.} We  evaluate the effects of DMD's key components, including feature decoupling (FD), HomoGD, crossmodal attention unit (CA), HeteroGD. The results are illustrated in Tab.~\ref{tab:Ablation}. We conclude the observations below.

\textbf{Firstly}, FD enhances MER performance significantly, it indicates the decoupled and refined features can reduce information redundancy and provide discriminative multimodal features.
To further prove the effectiveness of FD, we conduct experiments on our baseline model with and without FD on MOSEI dataset. As shown in Tab.~\ref{tab:fd}, FD brings consistent improvements for each unimodality. Meantime, the performance gap for the three modalities is reduced as the standard deviations of ACC$_{2}$ and F1 are both decreased.
\textbf{Secondly}, combing FD with HomoGD brings further benefits. 
Although the homogeneous features are embedded in the same-dimension space, there still exists different discriminative capabilities for the modalities.
HomoGD can improve the weak modalities through GD. To verify this, we conduct experiments with or without HomoGD on MOSEI dataset. The ACC$_2$ results are:  \underline{80.9\%} vs. \underline{82.4\%} for \textit{language}, \underline{56.5\%} vs. \underline{61.8\%} for \textit{vision}, \underline{54.4\%} vs. \underline{64.1\%} for \textit{audio}.
However, conducting HeteroGD without the crossmodal attention units will generate degraded performance, indicating the multimodal transformer plays a key role in bridging the multimodal distribution gap. \textbf{Thirdly}, with CA units and HeteroGD, DMD obtains conspicuous improvements, suggesting the importance of the taking advantage of the modality-exclusive features for robust MER.

Besides, we compare our proposed DMD with the classical MulT~\cite{MulT} for further investigation.  The results are shown in Tab.~\ref{tab:mult with gd}, where MulT (\textit{w/ GD}) means we add a GD-Unit on MulT to conduct adaptive knowledge with the reinforced multimodal features. Essentially, the core difference between MulT (\textit{w/ GD}) and DMD is that DMD incorporates feature decoupling. The quantitative comparison in Tab.~\ref{tab:mult with gd} shows that DMD obtains consistent improvements than MulT (\textit{w/ GD}). It suggests decoupling the multimodal features before distillation is feasible and reasonable. Furthermore, DMD achieves more significant improvements than the vanilla MulT, indicating the benefits of combing the feature decoupling and the graph distillation mechanisms.

\textbf{Visualization of the decoupled features.}
We visualize the decoupled homogeneous and heterogeneous features of DMD, DMD (\textit{w/o Hom., Het.}), DMD (\textit{w/o Het.}) in Fig.\ref{fig:tsne-c} and Fig.\ref{fig:tsne-s} for a quantitative comparison. 
DMD (\textit{w/o Hom., Het.}) denotes DMD without HomoGD and HeteroGD.
Besides, DMD (\textit{w/o Het.}) means DMD without HeteroGD. 
To visualize the homogeneous features, we randomly select 28 samples (four samples for each emotion category) in the testing set of the CMU-MOSEI dataset
For the heterogeneous features, we randomly select 400 samples in the testing set of the CMU-MOSEI dataset.
The features of the selected samples are projected into a 2D space by t-SNE.
 
For the homogeneous multimodal features of DMD and DMD (\textit{w/o Het.}), the samples belonging to the same emotion category naturally cluster together due to their inter-modal homogeneity.
With the decoupled homogeneous features but without graph distillation mechanism in DMD (\textit{w/o Hom., Het.}), the samples merely show basic separability for the binary \textit{non-negative} and \textit{negative} categories. However, the samples are not distinguishable under the 7-class setting, indicating the features are not so discriminative than that of DMD or DMD (\textit{w/o Het.}). The comparison between DMD (\textit{w/o Hom., Het.}) and DMD, and the comparison between DMD (\textit{w/o Hom., Het.}) and DMD (\textit{w/o Het.}) verifies the effectiveness of the graph distillation on the homogeneous multimodal features.

In the heterogeneous space, due to its inter-modal heterogeneity, the features of different samples should cluster by modalities.
As shown in Fig~\ref{fig:tsne-s}, DMD shows the best feature separability, indicating the complementarity between modalities is mostly enhanced.
DMD (\textit{w/o Hom., Het.}) and DMD (\textit{w/o Het.}) show less feature separability than DMD, indicating the importance of the graph distillation on the heterogeneous multimodal features.

\textbf{Visualization of graph edges in the GD-Units.} 
We show the dynamic edges in each GD-Unit in Fig.~\ref{fig:edge} for analysis. Each graph edge corresponds to the strength of a directed distillation.
We conclude the observations as: (1) The distillation in HomoGD are mainly dominated by $L \to A$ and $L \to V$. This is because the decoupled homogeneous language modality still plays the most critical role and outperforms visual or acoustic modality with significant advantages. For binary MER on the CMU-MOSEI dataset,  \textit{language}, \textit{visual}, \textit{acoustic} modality respectively obtains 80.9\%, 56.5\%, 54.4\% accuracy using the decoupled homogeneous features.  
(2) For HeteroGD,  $L \to A$, $L \to V$, and $V \to A$ are dominated. An interesting phenomenon is that $V \to A$ emerges.
This should be reasonable because the \textit{visual} modality enhanced its feature discriminability via the multimodal transformer mechanism in HeteroGD.
Actually, the three modalities obtain 84.5\%, 83.8\%, 71.0\% accuracy, respectively. Conclusively, the graph edges learn meaningful patterns for adaptive crossmodal distillation.


\section{Conclusion and discussion}
Within this paper we have proposed a decoupled multimodal distillation method (DMD) for MER. 
Our method is inspired by the observation that the contribution of different modalities varies significantly. Therefore, robust MER can be achieved by distilling the reliable and generalizable knowledge across the modalities. To mitigate the modality heterogeneities, DMD decouples the modal features into modality-irrelevant/-exclusive spaces in a self-regression manner. Two GD-Units are incorporated for each decoupled features to facilitate adaptive cross-modal  distillation. Quantitative and qualitative experiments consistently demonstrate the effectiveness of DMD. A limitation is that DMD does not explicitly consider the intra-modal interactions. We will explore this in future work.

\textbf{Acknowledgement:} This work was supported by the National Natural Science Foundation of China (62102180, 62072244), the Natural Science Foundation of Jiangsu Province (BK20210329), Shuangchuang Program of Jiangsu Province(JSSCBS20210210).


{\small
\bibliographystyle{ieee_fullname}
\bibliography{egbib}
}

\end{document}